\documentclass[sigconf,nonacm]{acmart}

\usepackage[utf8]{inputenc}

\usepackage{url}

\usepackage{makecell}

\usepackage{csquotes}

\usepackage{graphicx}
\graphicspath{{images/}}

\usepackage{pgfplots}
\pgfplotsset{compat=1.17}

\usepackage{multirow}

\usepackage{makecell}

\usepackage{array}

\usepackage{enumitem}

\usepackage{lipsum}

\usepackage{fontawesome}

\usepackage{multibib}
\newcites{P}{Primary Studies}

\usepackage{pifont}

\newcommand{\review}[1]{{#1}}

\newcommand{\luis}[1]{\textcolor{teal}{\ding{46}~\textbf{Luis:~}#1}}

\usepackage{xspace}
\newcommand{\ie}{\emph{i.e.,}\xspace}
\newcommand{\eg}{\emph{e.g.,}\xspace}

\newcommand{\etal}{\emph{et~al.}\xspace}

\newcommand{\ga}{Green AI\xspace}

\newenvironment{nbquote}
 {\quote\interlinepenalty=10000 }
 {\endquote}

\usepackage{framed}
\definecolor{formalshadelight}{RGB}{242,242,242}
\definecolor{formalshadedark}{RGB}{166,166,166}
\newenvironment{highlight}{%
  \MakeFramed{\advance\hsize-\width\FrameRestore}%
  \noindent\begin{minipage}{\linewidth}\noindent\hspace{-4.55pt}
  \vspace{2pt}\vspace{2pt}
}
{%
  \vspace{2pt}\vspace{-2pt}\end{minipage}\endMakeFramed%
}

\usepackage{tcolorbox}

\usepackage{listings}
\usepackage{balance}

\usepackage[compact]{titlesec}
\titlespacing*{\section}{0pt}{5pt}{4pt}
\titlespacing*{\subsection}{0pt}{4pt}{4pt}
\titlespacing*{\subsubsection}{0pt}{2pt}{2pt}

\AtBeginDocument{%
  \providecommand\BibTeX{{%
    \normalfont B\kern-0.5em{\scshape i\kern-0.25em b}\kern-0.8em\TeX}}}

\setcopyright{acmcopyright}
\copyrightyear{2021}
\acmYear{2021}
\acmDOI{10.1145/1122445.1122456}

\usepackage{fontawesome}
\newcommand{\papercount}[1]{\textit{$\triangleright$~\textbf{#1} out of 98 papers.}}

\begin{document}

\title{A Systematic Review of Green AI}

\author{Roberto Verdecchia}
\affiliation{
  \institution{University of Florence}
  \city{Florence}
  \country{Italy}
}
\email{roberto.verdecchia@unifi.it}

\author{June Sallou}
\affiliation{
  \institution{TU Delft}
  \city{Delft}
  \country{The Netherlands}
}
\email{j.sallou@tudelft.nl}

\author{Lu\'{i}s Cruz}
\affiliation{
  \institution{TU Delft}
  \city{Delft}
  \country{The Netherlands}
}
\email{l.cruz@tudelft.nl}

\begin{abstract}

With the ever-growing adoption of AI-based systems, the carbon footprint of AI is no longer negligible. AI researchers and practitioners are therefore urged to hold themselves accountable for the carbon emissions of the AI models they design and use. This led in recent years to the appearance of researches tackling AI environmental sustainability, a field referred to as Green AI. Despite the rapid growth of interest in the topic, a comprehensive overview of Green AI research is to date still missing. To address this gap, in this paper, we present a systematic review of the Green AI literature. From the analysis of 98 primary studies, different patterns emerge. The topic experienced a considerable growth from 2020 onward. Most studies consider monitoring AI model footprint, tuning hyperparameters to improve model sustainability, or benchmarking models. A mix of position papers, observational studies, and solution papers are present. Most papers focus on the training phase, are algorithm-agnostic or study neural networks, and use image data. Laboratory experiments are the most common research strategy. Reported Green AI energy savings go up to 115\%, with savings over 50\% being rather common. Industrial parties are involved in Green AI studies, albeit most target academic readers. Green AI tool provisioning is scarce. As a conclusion, the Green AI research field results to have reached a considerable level of maturity. Therefore, from this review emerges that the time is suitable to adopt other Green AI research strategies, and port the numerous promising academic results to industrial practice.

\end{abstract}


\maketitle

\section*{}
\begin{figure}[!h]
    \centering
    \includegraphics[width=\linewidth]{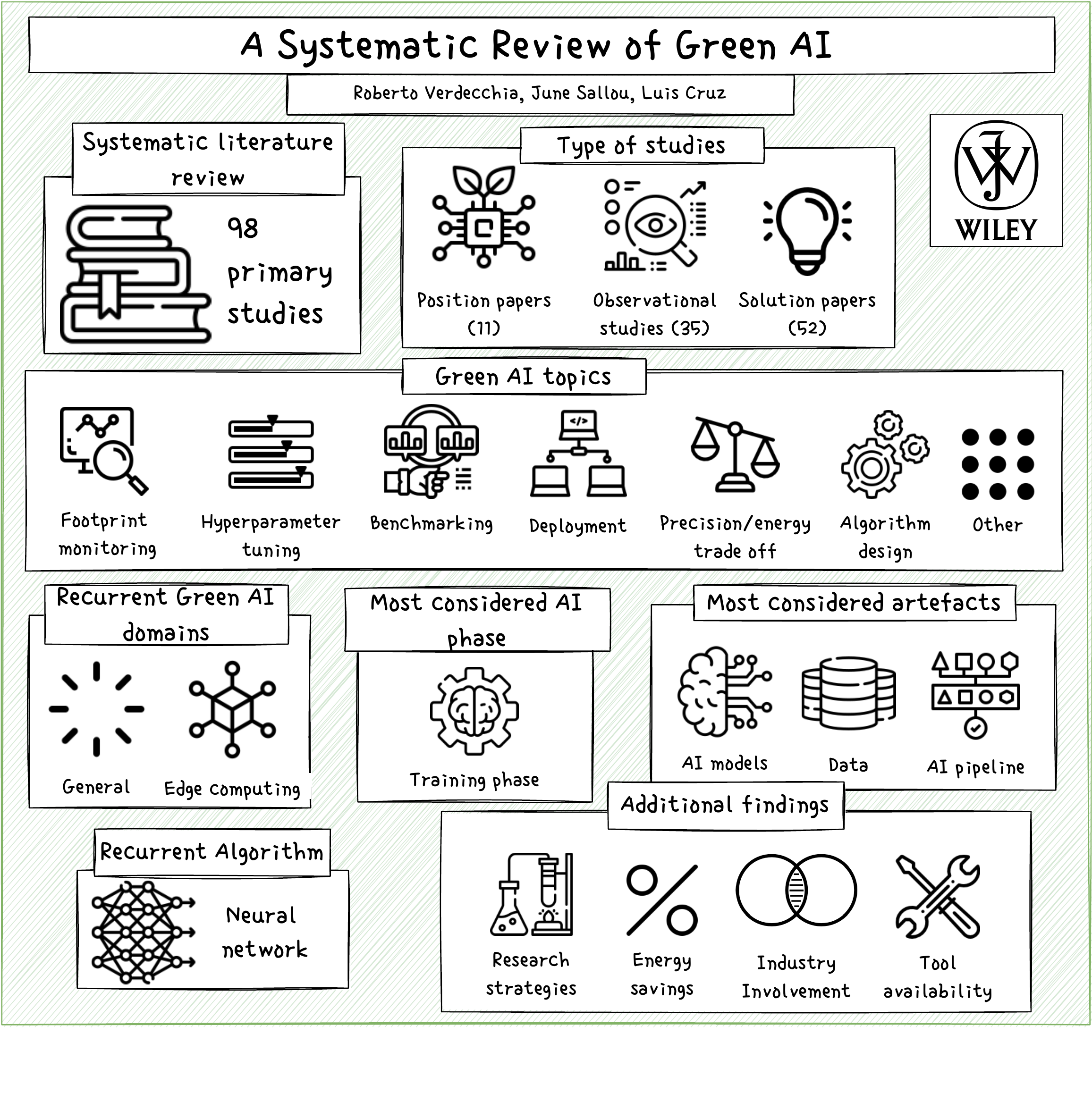}
    \caption*{Graphical Abstract: From a systematic review of the Green AI literature, Green AI results to focus on solutions, and is often not bound to a specific context or algorithm. The Green AI research field results to be mature, \ie the moment is suitable to port results from academic research to industrial practice.}
    \label{fig:graph_abstract}
\end{figure}

\section{Introduction}
\label{sec:intro}

In recent years, the Artificial Intelligence (AI) community has been challenged to bring the carbon footprint of AI models to the top of their research agenda.
The iconic paper by \citeauthor{Strubell2019Monitoring} ~\citeP{Strubell2019Monitoring} analyzes the carbon impact of training their own state-of-the-art models. Results lead to the conclusion that we need to reduce the carbon footprint of developing and running AI models.

This self-reflection was an eye-opener to the AI research community.
Many papers followed, calling for a new research direction that would consider this problem. \citeauthor{Schwartz2020GreenAI} coined the term Green AI as \textit{``AI research that yields novel results while taking into account the computational cost''}~\citeP{Schwartz2020GreenAI}. \citeauthor{Bender2021Parrots} published a position paper highlighting the consequences of continuously increasing the size of AI models~\citeP{Bender2021Parrots}. A natural question that is posed is whether we are doing enough as a research community to mitigate the carbon impact of developing and running AI-based software.

AI systems are significantly complex and, to achieve Green AI, we need a joint effort that targets all the different stages of an AI system's lifecycle (e.g, data collection, training, monitoring), different artifacts (\eg data, model, pipeline, architecture, hardware), etc~\cite{haakman2021ai}. 

Given the heterogeneity of the field, it is also difficult to have a broad view of all the Green AI literature that has been published in the past years.
To understand the existing research, we conduct a systematic literature review on Green AI. We provide an overview and characterization of the existing research in this field. Moreover, we study how the field has been evolving over the years, pinpoint the main topics, approaches, artifacts, and so on. 

This literature review shows that there has been a significant growth in Green AI publications -- 76\% of the papers have been published since 2020.
The most popular topics revolve around monitoring, hyperparameter tuning, deployment, and model benchmarking. We also highlight other emerging topics that might lead to interesting solutions -- namely, \textit{Data Centric} Green AI, \textit{Precision/Energy Trade-off} analysis.
The current body of research has already showcased promising results with energy savings from 13\% up to 115\%. Still, most of the existing work focuses on the training stage of the AI model.
Moreover, we observe that there is little involvement of the industry (23\%) and that most studies revolve around laboratory experiments. We argue that the field is growing to a level of maturity in which involvement of the industry is quintessential to enable the overarching goal of \ga: harness the full potential of AI without a negative impact in our planet. 
 
To encourage open science and the reproducibility of this study, we provide all data and scripts in a replication package available online with an open-source license\footnote{Replication package: \url{https://github.com/luiscruz/slr-green-ai}}.

The remainder of this paper is structured as follows. In Section~\ref{sec:method}, we describe the methodology used to collect and analyze Green AI literature. In Section~\ref{sec:results}, we present all the results yielded by our methodology. Section~\ref{sec:discussion} discusses findings and reflects on the impact of our results in the research community. In Section~\ref{sec:threats}, we reflect on the potential threats to the validity of this study. Following, Section~\ref{sec:rw} describes related work and pinpoints the differences with our study. The main conclusions and future work are presented in Section~\ref{sec:conclusion}.


\begin{figure*}
    \centering
    \includegraphics[width=1\textwidth]{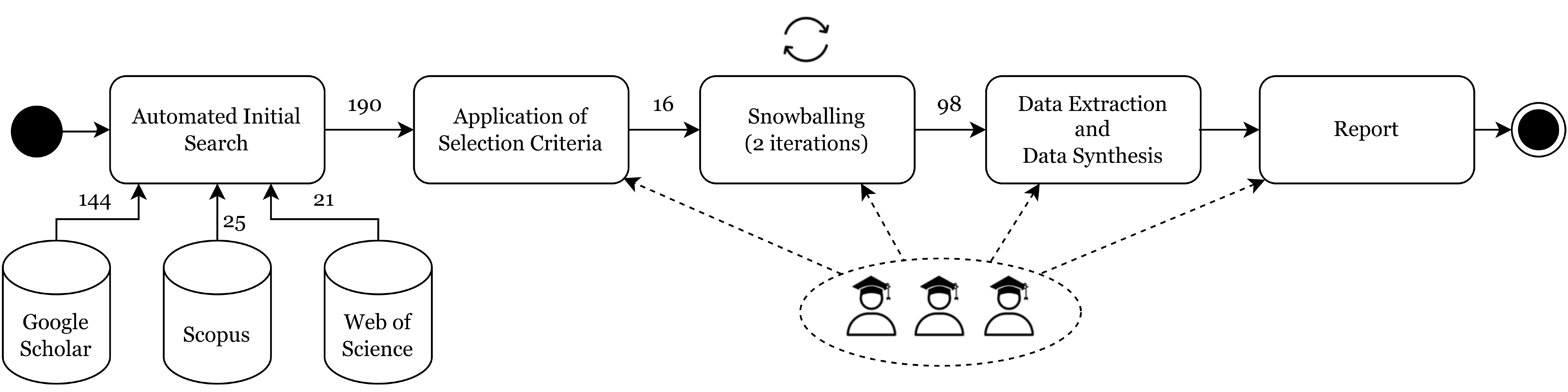}
    \caption{Systematic literature review process overview.}
    \label{fig:process}
\end{figure*}

\section{Methodology}
\label{sec:method}
In this section, we document the research design, which was rigorously adhered to during the planning and execution of the study.
We primarily followed the guidelines for conducting SLRs in software engineering research presented by Kitchenham~\cite{Kitchenham2004}.

\subsection{Research Objective and Question}
\label{sec:objAndRq}
The goal of this review is to understand the characteristics of existing \ga research.
By utilizing the Goal-Question-Metric method~\cite{gqm}, this objective can be described more formally as follows:
\vspace{3pt}

\noindent\textit{
    \textbf{Analyze} \ga literature\\
    \textbf{For the purpose of} knowledge collection and categorization\\
    \textbf{With respect to} AI \\
    \textbf{From the viewpoint of} researchers and practitioners\\
    \textbf{In the context of} environmental sustainability.
}

\vspace{3pt}
The goal of this research can be directly translated into a research question (RQ), which states as follows:

\begin{enumerate}[label=\textbf{RQ:},leftmargin=*]
    \item \textit{What are the characteristics of \ga state-of-the-art research?}

\end{enumerate}
By answering our research question, we aim at gaining a systematic overview of the \ga body of knowledge, starting from an outline of the general publication trends, to a detailed analysis of the past and current \ga research activities and their characteristics.

\subsection{Research Process}
An overview of the research process followed is depicted in Figure~\ref{fig:process}.
The process starts with the execution of a conservative automated search query via the digital libraries and indexing platforms \textit{Google Scholar}, \textit{Scopus}, and \textit{Web of Science},
complemented by a subsequent iterative bidirectional snowballing process, which is conducted until the achievement of theoretical saturation.
Including multiple literature indexing platforms to execute the automated search allows us to conduct an encompassing search of the literature based on multiple sources, hence allowing us to mitigate potential threats to external validity, as further documented in Section~\ref{sec:threats}. Following, the details of each step of our research process are documented in~detail.

\subsubsection{Automated Initial Search.}
\label{sec:automated_query}
To identify a preliminary set of potentially relevant research works, we design an encompassing automated query to be executed on three different literature indexers, namely \textit{Google Scholar}, \textit{Scopus}, and \textit{Web of Science}. The automated query targeting publication titles states as follows:

\vspace{-20pt}
\begin{lstlisting}[escapechar=@,  language=sql, caption={Automated search query}, label={list:query}, numbers=left, abovecaptionskip=3em, basicstyle=\small	\ttfamily, abovecaptionskip=3em]
INTITLE("green" OR "sustainab*") AND
INTITLE("AI" OR "ML" OR "artificial intelligence" 
OR "machine learning" OR "deep learning")
\end{lstlisting}

The query is designed to retrieve literature with titles containing keywords related to sustainability, identified by the keywords \textit{green} or \textit{sustainability} and its variations, \eg ``sustainable''  (Listing~\ref{list:query}, Lines 1). The second part of the query instead is used to retrieve literature concerning AI, or related synonyms and acronyms (Listing~\ref{list:query}, Lines 2-3).
The query is executed on the three aforementioned literature libraries and indexes on the 18th of July 2022, and led to the identification of 190 potentially relevant studies. In order to be as comprehensive as possible, and avoid potential threats to external validity, the year of publication is left unbounded in the automated search. 

\subsubsection{Application of Selection Criteria.}
Subsequent to the identification of the initial potentially relevant studies, we execute the manual selection of the studies via a set of selection criteria defined \textit{a priori}. A paper is confirmed as primary study if it adheres to all inclusion criteria, and none of the exclusion ones.
The following inclusion (I) and exclusion (E) criteria are used:

\begin{itemize}
    \item[I1-] The study regards AI
    \item[I2-] The study regards environmental sustainability
    \item[I3-] The study regards the environmental sustainability of AI
    \item[I4-] The study regards the software level 
    \item[E1-] The study is not written in English
    \item[E2-] The study is not available    
    \item[E3-] The study is a duplicate or extensions of an already included study
    \item[E4-] The study is a secondary or tertiary study
    \item[E5-] The study is in the form of editorials, tutorials, books, extended abstracts, etc.
    \item[E6-] The study is a non-scientific publication or grey literature
\end{itemize}

With the first three inclusion criteria (I1-I3), we ensure that the primary studies focus on \ga (I1, I2), and that the studies regard the environmental sustainability \textit{of} AI, rather than the improvement of environmental sustainability \textit{through} AI. With the fourth inclusion criterion instead (I4), we ensure that the primary studies focus on software-centric \ga. This latter criterion is used to exclude studies focusing on hardware-specific \ga techniques, \eg the use of \textit{ad hoc} implemented hardware components, which we consider out of reach for most researchers/practitioners interested in \ga, and is only marginal to the definition of \ga itself~\cite{Schwartz2020GreenAI}.

The exclusion criteria are designed to ensure that data can be extracted from the papers (E1, E2), do not represent duplication or redundancy with respect to other primary studies (E3, E4), and are provided in the form of scientific studies (E5, E6).

To ease the primary study selection process, adaptive reading depth~\cite{Petersen2008} is used to efficiently assess potentially relevant studies.
In order to mitigate subjective biases and interpretations, the three authors independently utilized the selection criteria to scrutinize 63-64 candidate studies. Weekly meeting are held during the selection process to jointly discuss examples, doubts, and align the selection process between the three researchers.

The application of the selection criteria concludes with the identification of 16 primary studies, which constitute the starting set for the subsequent snowballing process.

\subsubsection{Snowballing.}
In order to enrich the set of selected primary studies, and ensure that the primary study comprehensively represents the \ga body of literature, the automated search results are complemented with a recursive bidirectional snowballing process~\cite{Wohlin2014}.
This step entails the scrutiny of all studies either citing or cited by the already included primary studies.
As for the application of selection criteria, three researchers are involved in the snowballing. During each snowballing round, the researchers independently snowball different primary studies, and propose new primary studies to be included, \ie the new identified studies which adhere to the selection criteria. During each snowballing round, examples, doubts, and divergences are jointly revisited and resolved, and the next snowballing iteration is started.
A total of two rounds of backward and forward snowballing are executed before no new studies are identified, \ie when theoretical saturation is reached. 
The snowballing process terminates with the inclusion of 82 new primaries studies, leading to a total of 98 primary studies which are considered in the literature review reported in this research.

\subsubsection{Data Extraction.}
\label{sec:data_extraction}
In order to achieve the intended goal of this study and answer our RQ (see Section~\ref{sec:objAndRq}), we proceed to systematically extract data from the primary studies. The data extraction process consisted of two subsequent phases. 

The first phase consists of a data exploration process, which terminates with the establishment of the data extraction framework of this study. Specifically, during this first phase, the three authors of this review independently scan the identified primary studies, and annotate the characteristics of the studies which are relevant to answer our RQ. The identified characteristics are then jointly discussed and refined, leading to the consolidation of the fields constituting the data extraction framework of this review.

In the second data extraction phase, the primary studies are thoroughly analyzed, and the data is extracted from the studies according to the data extraction framework. 

The fields of the data extraction framework utilized for this literature review on \ga are the following.

\begin{itemize}
    \item \review{\textit{Green AI Definition:} the level of abstraction used in the paper to quantify the impact of AI in the surrounding environment: energy efficiency~\cite{verdecchia2021green}, carbon footprint~\cite{wiedmann2008definition}, or ecological footprint~\cite{matuvstik2021footprint}.} 
    \item \textit{Study type:} The overarching type of study, which could be either presenting a position on \ga, a \ga solution, or an observational study on \ga;
    \item \textit{Topic:} The \ga topic considered in the study, \eg hyperparameter-tuning to achieve energy efficiency of an AI algorithm;
    \item \textit{Domain:} The domain considered in the study, \eg edge or mobile computing; 
    \item \textit{Type of data:} The type of data utilized by AI in the study, \eg text or images;
    \item \textit{Artifact considered:} The AI artifact considered in the study, \eg the data used by AI models, the AI models themselves, or the AI deployment pipeline.
    \item \textit{Considered phase:} If the study focused on the AI training phase, the AI inference phase, or both.
    \item \textit{Research strategy:} The research strategy, as defined in ~\cite{stol2018abc}, used to support the claims reported in the study;
    \item \textit{Dataset size:} The size of the dataset, in number of data points, considered in the study (if any);
    \item \textit{Energy Savings:} The reported percentage energy savings achieved by solutions reported in the study (if any is documented);
    \item \textit{Industry involvement:} Industry involvement in the authorship of the study, which could be either academic-only authorship, industrial-only authorship, or mixed authorship;
    \item \textit{Intended reader:} If the study is primarily intended for academic readers, industrial readers, or the general public.
    \item \textit{Tool availability:} The availability of the tool(s) to address \ga presented in the study (if any).
\end{itemize}


\subsubsection{Data Synthesis.}
During the data extraction process, the data was harmonized by relying on the constant comparison~\cite{glaser1965constant} of extracted keywords, breaking up keywords into more specific ones when their semantic depth required it, or merging very similar keywords to avoid redundancy. This analysis process relied on open coding~\cite{jenner2004companion} to systematically identify recurrent concepts, followed by axial coding~\cite{jenner2004companion} to manage the increasing complexity of some emerging concepts.

The only exceptions were made for the \textit{research strategy}, \textit{industry involvement}, and \textit{tool availability} fields of the extraction framework (see Section~\ref{sec:data_extraction}), for which provisional coding was used~\cite{jenner2004companion}. Specifically, coding of the \textit{research strategy} relied on the research strategy categories reported by Stol~\etal~\cite{stol2018abc} was used. The industry involvement instead relied on three pre-defined fields, namely ``academic-only authorship'', ``industrial-only authorship'', or ``mixed authorship''. Finally, \textit{tool availability} could only assume one of two pre-defined values, namely ``Yes'' (if the tool is available) or ``No'' (if the tool is not available, or none is presented in the primary study).

During the data extraction and synthesis phase, emerging codes are continuously discussed among the three authors of the review. This process ensures that the emerging codes and their abstraction level are kept consistent among researchers, and are aligned with the research goal and question of the study.

\section{Results}
\label{sec:results}
In this section, we present the results collected with our SLR on Green AI.

\subsection{Publication Years}
\begin{figure}[!h]
    \centering
    \includegraphics[width=\linewidth]{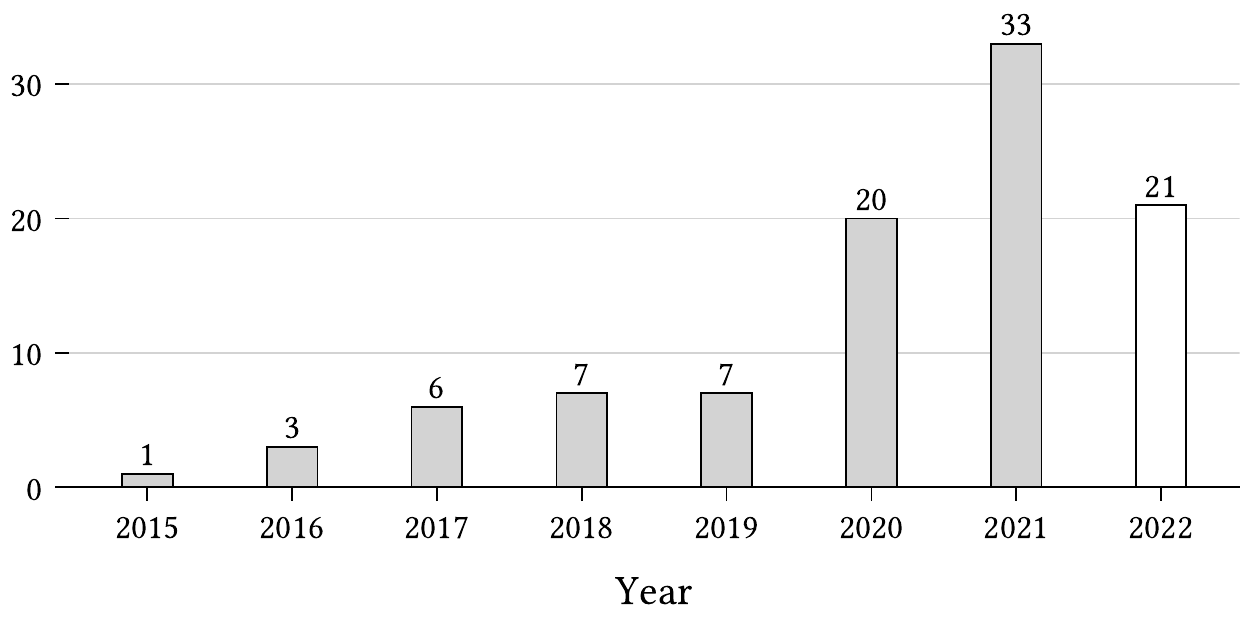}
    \caption{Number of publications per year.}
    \label{fig:res_year}
\end{figure}

The literature spans from 2015 with the first publication on the topic to this present year (\ie 2022). Figure~\ref{fig:res_year} presents the distribution of the literature papers regarding the publication year. We observe a global increase following the years. Furthermore, a spike in the number of publications is seen in 2020, going from 7 publications in 2019 to 20 in 2020. 
As the automated initial search was launched in 2022, the publication trends reported in this review might not be representative of the actual research output of 2022 (see also Section~\ref{sec:automated_query}). 

\subsection{Venue Types}
Publications are particularly concentrated on conferences (\papercount{47}) and journals (\papercount{39}). Only 12 out of the 98 publications are associated with a workshop. Conferences being treated as an equal publishing venue as journals follows the trends observed in the computer science research field \cite{Vrettas2015, Kim2019}.

\begin{highlight}
\textbf{Green AI publication trends}\newline
\noindent \faLeaf~The topic of \ga is experiencing an increasing trend of popularity, with a considerable growth in publications from 2020 onward. Most studies are published in conferences and journals, while only a minor portion in workshops.
\end{highlight}

\subsection{Green AI Definition}
\label{sec:green_ai_definition}
\review{%
The distribution of publications across different \ga definitions is presented in Figure~\ref{fig:green_ai_definition}. Most literature addresses \ga at the level of energy efficiency (81 papers). Higher-level definitions, namely carbon and ecological footprint, are only addressed in 20 and 9 publications respectively. Note that a primary study might be mapped to more than one definition, if more than one is used in the paper at hand.}

\noindent{\begin{figure}
    \centering
    \includegraphics[width=0.85\linewidth]{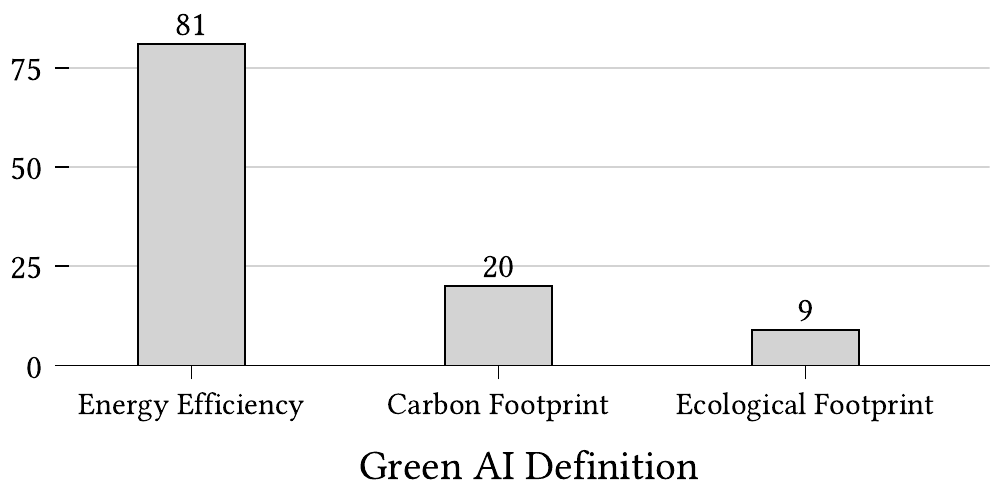}
    \caption{Number of publications per type of Green AI definition.\label{fig:green_ai_definition}}
\end{figure}

\subsection{Study Types}
\label{sec:study_type}
Existing literature on Green AI spans across three types of studies, namely \textit{observational}, \textit{solution}, and \textit{position} papers (see also Section~\ref{sec:data_extraction}). As shown in Figure~\ref{fig:study_type}, from the 98 papers covered in this review, the most common are solution papers, with 51 entries, followed by observational with 35, and position papers with 12. Note that study types are mutually exclusive, \ie a single paper has only one study type.

\noindent{\begin{figure}
    \centering
    \includegraphics[width=0.85\linewidth]{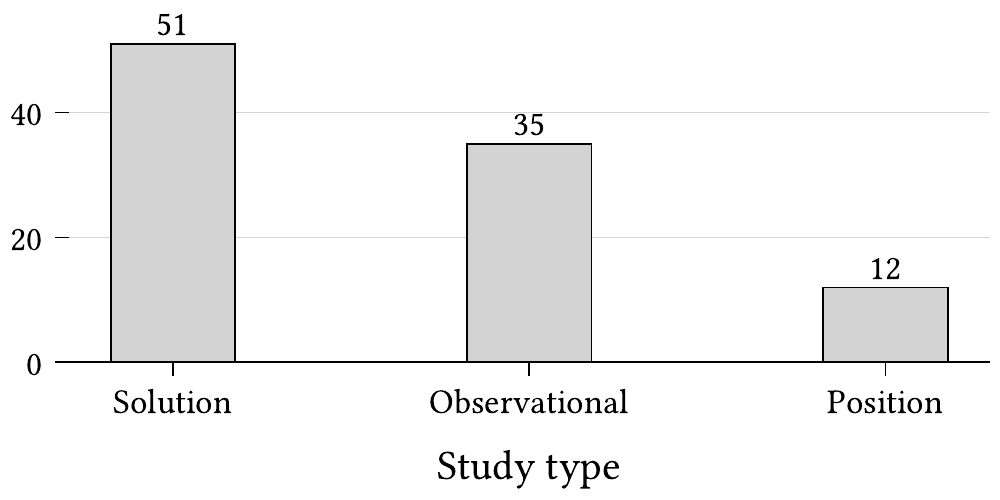}
    \caption{Number of publications per study type.\label{fig:study_type}}
\end{figure}

\subsection{\ga Topics}
\label{sec:topic}

From our analysis we identify 13 main topics being addressed by the Green AI literature. Figure~\ref{fig:topic} depicts the distribution of publications across the different topics.
The most popular topic is \textit{Monitoring}, addressed by 28 papers, followed by \textit{Hyperparameter Tuning} (18), \textit{Model Benchmarking} (17), \textit{Deployment} (17), and \textit{Model Comparison} (17).
Since papers are not exclusive to a single topic, these top-4 topics alone cover 61\% of the papers in this review.
Below, we pinpoint each topic with a short summary and the respective number of publications.

\paragraph{\textbf{Monitoring}} \papercount{28}
Covering monitoring approaches to study the energy and/or carbon footprint of AI models.
 
In this topic, papers report and reflect on the energy footprint of state-of-the-art models throughout their lifecycle. For example, Wu~\etal~\citeP{wu2022sustainable} provide a landscape of the carbon footprint of AI models across Facebook. Findings showcase that, typically, throughout the lifetime of AI models, 50\% of their carbon cost lies in the embodied carbon footprint of the hardware used to develop these models. However, the paper shows that the vast majority of training workflows under-utilizes GPUs at 30–50\% of their full capacity.

Other papers within this topic focus on solutions to make carbon monitoring feasible in any AI project~\citeP{garcia2019estimation}. As an example, the Carbontracker offers a toolset to track and predict the energy and carbon footprint of training DL models~\citeP{anthony2020carbontracker}. These studies argue that it is quintessential to report the energy and carbon footprint of model development and training alongside performance metrics.
 
\paragraph{\textbf{Hyperparameter Tuning}} \papercount{18}
Improving or assessing the impact on the energy consumption of optimizing hyperparameters when training an AI model.

Many publications are motivated by the fact that tuning parameters leads to significant energy costs -- it requires retraining a model multiple times in order to find the optimal set of hyperparameter values. Hence, most publications within this topic focus on identifying alternative strategies that reduce the number of iterations required to tune hyperparameters~\citeP{Stamoulis2018}.

On a different perspective, Chavannes~\etal~\citeP{Puvis2021Hyperparameter} explore how hyperparameter tuning can help deliver more energy-efficient models by adding power consumption to the set of parameters being optimized.

\paragraph{\textbf{Model Benchmarking.}} \papercount{17}
Studies that contribute with benchmarks to compare the energy footprint of different models or training techniques. 

Benchmarks help the community understand how the state of the art behaves w.r.t. given performance indicators. Ultimately, they help create baselines so that new approaches can be properly validated and compared to the state of the art. As example of publications within this category, \citeauthor{Asperti2021} ~\citeP{Asperti2021} evaluate the energy cost of different variational autoencoders. Another study, by Yu~\etal~\citeP{Yu2022Jan}, compares the energy efficiency of common machine learning algorithms when applied to clinical laboratorial datasets.
  
 \paragraph{\textbf{Deployment}} \papercount{17}
 Addressing the deployment stage of the lifecycle of an AI model.

 Typically, publications in this topic discuss the problem of deploying AI models in a real scenario or in a scenario with peculiar constraints that challenge a standard approach. For example, deployment publications showcase the challenges of deploying energy-efficient AI in FPGA~\citeP{Tao2020}, in Edge devices~\citeP{Gondi2021Nov,kim2020}, in mobile devices~\citeP{Wang2022Federated,manasi2020,jayakodi2020}, and so on.  
 
 \paragraph{\textbf{Precision/Energy Trade Off}} \papercount{11}
 There is a turning point where to increase a very small fraction of the model performance, it is required to endure an energy-intensive training loop. Within this topic, papers address the Pareto trade-off between having optimal accuracy and/or optimal energy efficiency.
 
Zhang~\etal~\citeP{Zhang2018} study how removing neurons from neural networks affects both accuracy and energy consumption. Results indicate that a good portion of neurons are redundant and can be removed to reduce energy consumption without a significant impact on accuracy. At the same time, it shows that there is a turning point where removing neurons improves energy efficiency but significantly reduces accuracy. Hence, the two parameters always need to be analyzed together. Other works opt for optimizing energy while keeping accuracy loss within a negligible margin~\citeP{Wang2020May}.
 
\paragraph{\textbf{Algorithm Design}} \papercount{10}
Design of new training algorithms that produce models that are significantly more energy-efficient than the state of the art.

Some works propose small changes to the algorithms that make a big difference in the final energy consumption. For example, Garcia-Martin~\etal~\citeP{Garcia-Martin2021} approximate the splitting criteria by selecting branches that require less computational effort. Results showcase decision trees that are up to 31\% more energy efficient and with minimal impact on accuracy.
Other examples include Espnetv2~\citeP{mehta2019}, a lightweight convolutional neural network designed with power-efficiency in mind.

\paragraph{\textbf{Libraries}} \papercount{8}
Our choice of libraries have an impact on the final carbon footprint of AI systems. Studies within this topic provide some sort of evaluation of different AI libraries and how they contribute to energy efficiency. 

This category shows that software engineering studies play an important role in enabling Green AI. Georgiou~\etal~\citeP{Georgiou2022May} compare the energy footprint of deep learning frameworks. Results showcase that PyTorch is more energy-efficient than Tensorflow at the training stage. However, Tensorflow tends to be more energy-efficient at the inference stage. The study delves into the framework's different API methods and highlights code in the frameworks that should be optimized to reduce energy consumption. Finally, the authors motivate the importance of reporting and discussing energy efficiency in the documentation of deep learning frameworks.
  
\paragraph{\textbf{Data Centric}} \papercount{6}
Typically, the AI community has looked into coming up with better model training strategies. However, there is a new trend in AI that is raising the importance of developing better data collection and processing techniques as a more effective way to deliver better AI models. This line of thought within Green AI aims at reducing the carbon footprint of AI by tackling the problem at the data level.

Data-centric approaches for Green AI show that feature selection and subsampling techniques can significantly reduce the energy consumption of training machine learning models~\citeP{Verdecchia2022DataCentricGreenAI}. Subsampling strategies can be more sophisticated by removing data points that are expected to be redundant in terms of knowledge acquisition~\citeP{Dhabe2021}. 
 
\paragraph{\textbf{Network Architecture}} \papercount{6}
The impact of a distributed network on the energy efficiency of AI. AI models are often deployed in a distributed context -- \eg IoT, edge computing, etc. Hence the design and architecture of the network plays an important role in leveraging sustainable models.

For example, Kim and Wu~\citeP{kim2020} propose an adaptive execution engine that selects the inference strategy according to the signal strength of the network in different devices, as it is known to affect the energy efficiency of the edge mobile system.
  
 \paragraph{\textbf{Estimation}} \papercount{5}
 Collecting and making sense of energy or climate data is far from trivial -- many different factors contribute to the final estimation~\citeP{garcia2019estimation}. This topic revolves around understanding ways of estimating the energy consumption or carbon footprint of models.
 
 Existing solutions to estimate energy consumption for software fail to provide meaningful insight about energy consumption that can be mapped to a machine learning model's structure. IrEne creates a graph that breaks down NLP models into low-level machine learning primitives and provides energy estimations at the primitive level~\citeP{Cao}.
 
\paragraph{\textbf{Emissions}} \papercount{4}
Papers that focus on understanding the carbon impact of creating and/or consuming AI systems.
Dhar~\citeP{dhar2020} flags the importance of being able to quantify carbon impact and the lack of tools and data available. Fraga-Lamas~\etal~\citeP{Fraga-Lamas2021Aug} go beyond reporting the energy consumption of an AI-enabled IoT scenario and present how much carbon would be emitted in different countries and different energy sources.

 \paragraph{\textbf{Policy}} \papercount{3} Studies within this topic address and discuss strategies on how we should handle the carbon footprint of AI as a society.

Perucica and Andjelkovic~\citeP{Perucica2022Policy} reflect on the environmental policies implemented by the European Union, discussing whether they fit the AI era or new regulations are needed. Rhode~\etal~\citeP{rohde2021} call out for the unclear dilemma between the impact of existing/upcoming AI technologies and the commitment to achieve the 1.5\textcelsius{} climate change goal as expressed in the UNFCCC Paris Declaration.
 
\paragraph{\textbf{Ethics}} \papercount{3}
Papers that focus on the ethical implications of the growing carbon footprint of AI. Tamburrini~\citeP{Tamburrini2022Jan} discusses the responsibilities of AI scientists, AI infrastructure providers, and other stakeholders in enabling Green AI. The paper questions whether it is ethically justified to create massive AI pipelines to improve accuracy.
 
\paragraph{\textbf{Other}} \papercount{5}
Studies addressing a relevant topic with only a single publication in total: User values~\citeP{Konig2022}, Scheduling~\citeP{Zhu2021Aug}, Rebound Effects~\citeP{Willenbacher2021Rebound}, Security~\citeP{shumailov2021}, Energy Capping~\citeP{krzywaniak2022gpu}.

\begin{figure*}[!h]
    \centering
    \includegraphics[width=0.85\linewidth]{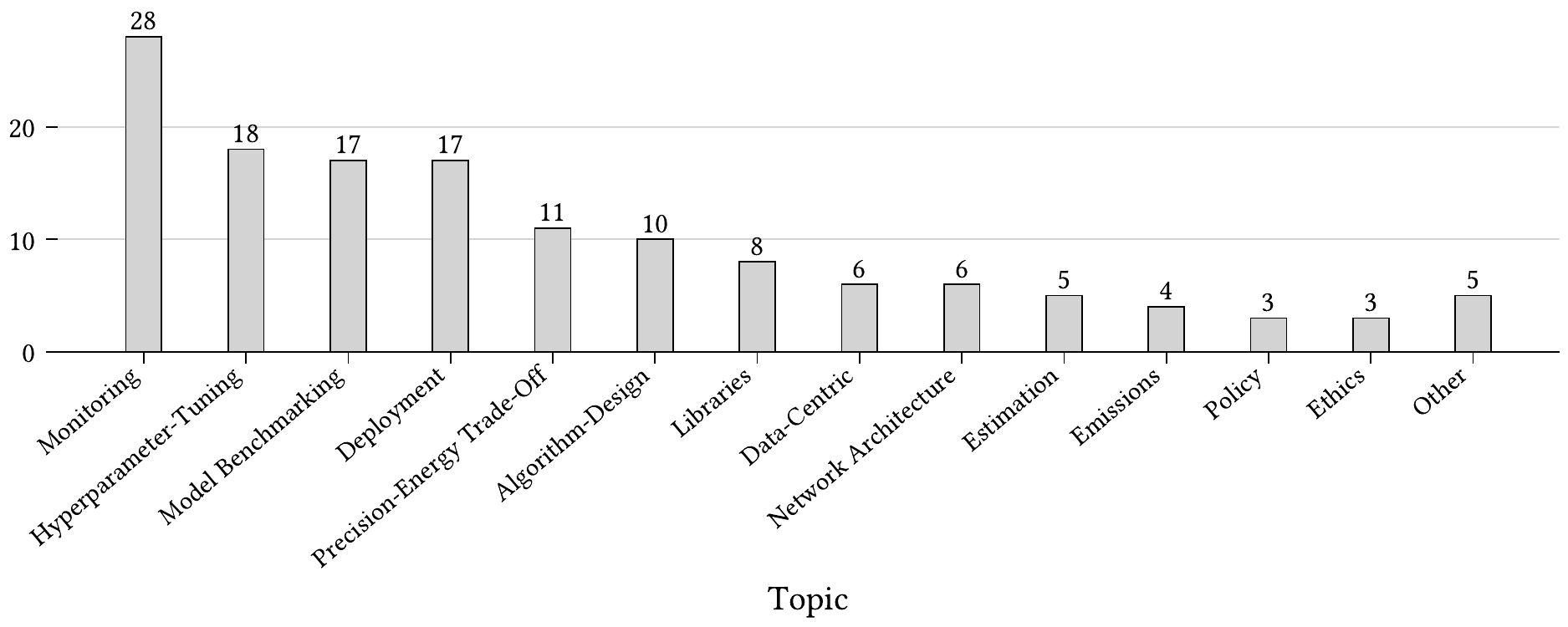}
    \caption{Number of papers per Green AI topic.}
    \label{fig:topic}
\end{figure*}

\begin{highlight}
\textbf{Green AI topics by study type}\newline
\noindent \faLeaf~There are 13 main topics on Green AI. The majority (61\%) of the publications focuses on Monitoring, Hyperparameter-tuning, Model Benchmarking, and Deployment. Despite being important, topics such as Data-Centric, Estimation, and Emissions are underrepresented in the scientific literature.
\end{highlight}

\subsection{Green AI Topics by Study Type}

We further investigate the distribution of papers across different topics per category. Figure~\ref{fig:bubble_topic} presents a bubble plot that draws a bubble for each pair topic (x-axis) and study type (y-axis). The size of the bubble is proportional to the number of papers published in each pair. The plot enables a few observations.

Most topics adhere to the general pattern observed earlier in Section~\ref{sec:study_type}: the majority of papers consist of solution studies, followed by observational and then position.
However, the topics of \textit{Model Benchmarking} and \textit{Libraries} do not follow this pattern, being mostly covered by observational papers. This is expected as these topics revolve around comparing different libraries and models to provide insight on the energy efficiency of different design decisions.

Moreover, papers from the least represented topics \textit{Ethics}, \textit{Policy}, and \textit{Emissions} tend to be position papers. From the ten studies in these three topics, only one is observational and none is solution.

Also worth noticing is the fact that the majority of the position studies in Green AI only cover the smallest topics. Considering the top-10 topics -- from Monitoring to Estimation -- only 6 are position papers. In contrast, the bottom-4 topics (including Other) are covered by 10 position papers.

\begin{figure}[!h]
    \centering
    \includegraphics[width=\linewidth]{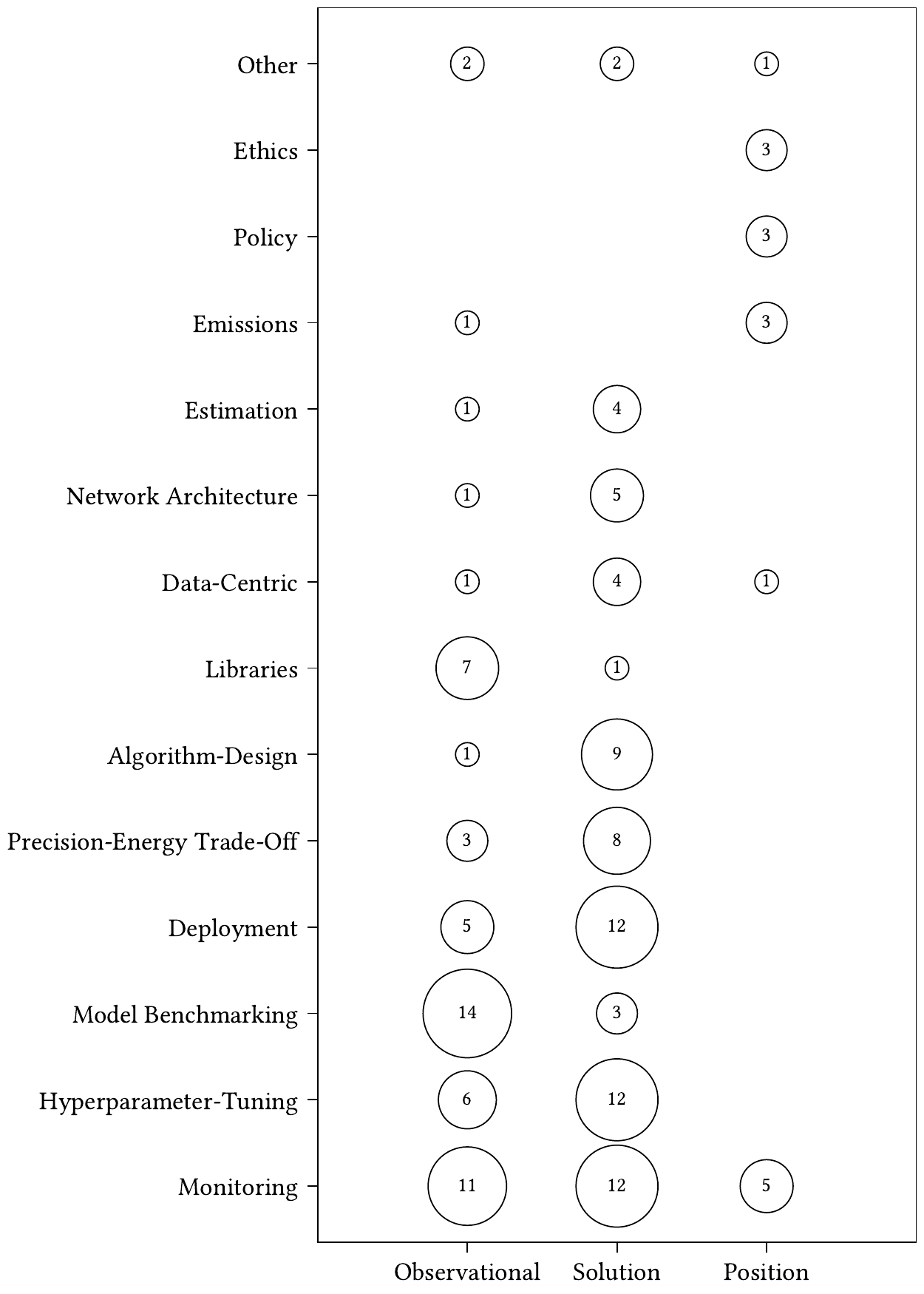}
    \caption{Number of publications by topic and study type.}
    \label{fig:bubble_topic}
\end{figure}

\begin{highlight}
\textbf{Green AI topics by study type}\newline
\noindent \faLeaf~Most publications on Ethics, Policy, and Emissions are position studies calling for more research in these topics.
\end{highlight}

\subsection{Domains}
Figure~\ref{fig:domain} presents the distribution of the publications according to the domain they cover.
The majority of the publications (\ie \papercount{58}) do not devote their studies to a specific domain, but tackle the energy efficiency of AI in a general context. 
Regarding the most specific studies, the most covered domains are:

\begin{description}
    \item [Edge] Regarding Internet of Things and Edge Computing, which are usually associated with distributed systems and networks. \papercount{24}
    \item [Computer Vision] Regarding image recognition.\papercount{6}
    \item [Cloud] \papercount{5}
    \item [Mobile] \papercount{4}
\end{description}

The \textbf{Other} category gathers publications about a specific domain, being covered only once, among Health, Autonomous Driving, Smart cities, Human Activity, Wearables, and Embedded Systems. 

\begin{figure}[!h]
    \centering
    \includegraphics[width=\linewidth]{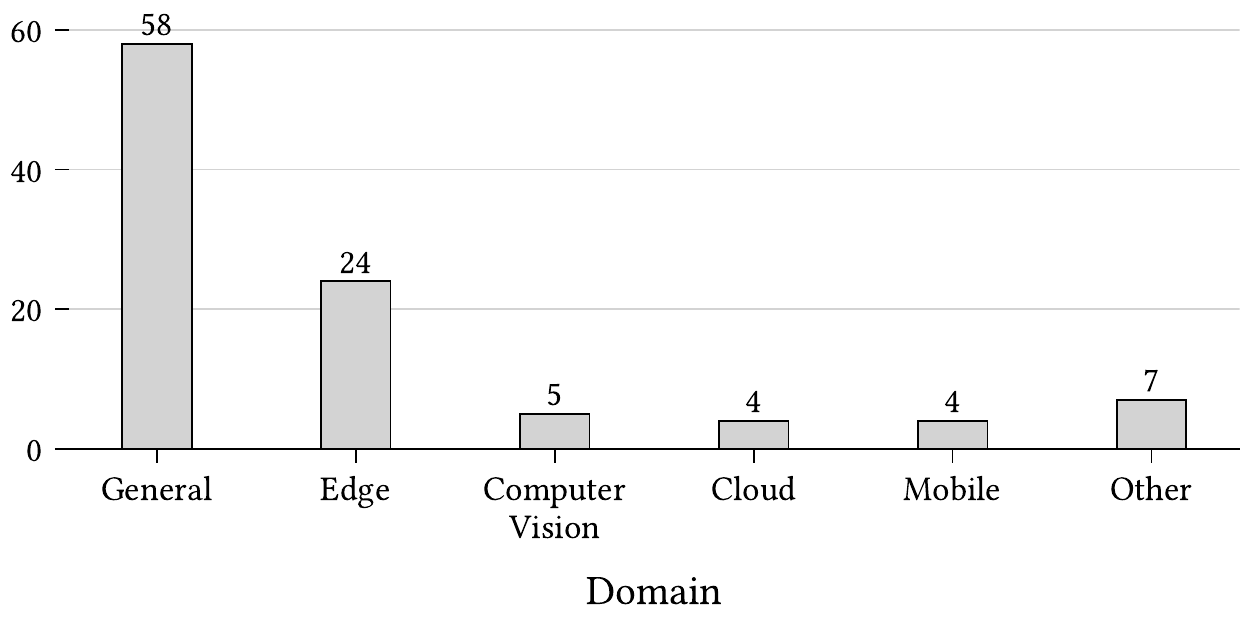}
    \caption{Number of publications per study domain.}
    \label{fig:domain}
\end{figure}

\begin{highlight}
\textbf{\ga domains}\newline
\noindent \faLeaf~ The majority of \ga studies does not focus a specific domain. Among specific domains, edge computing results to be the most recurrent one.
\end{highlight}

\subsection{AI Pipeline Phases}
The AI pipeline is divided into two major phases: the training, when the AI model is built, and the inference, when the model is used to make predictions from new data. Thus, we classify the papers according to 3 categories: \textbf{training}, \textbf{inference}, and \textbf{all}. The \textbf{all} category translates the fact that the paper does not consider a particular phase, but the whole pipeline.

As depicted in Figure~\ref{fig:res_phase}, we find that most of the publications on the topics of Green AI focus on the training phase (\papercount{49}). In comparison, fewer papers direct their studies at the inference phase (\papercount{17}) or on the overall process (\papercount{32}). 

\begin{figure}[!h]
    \centering
    \includegraphics[width=0.8\linewidth]{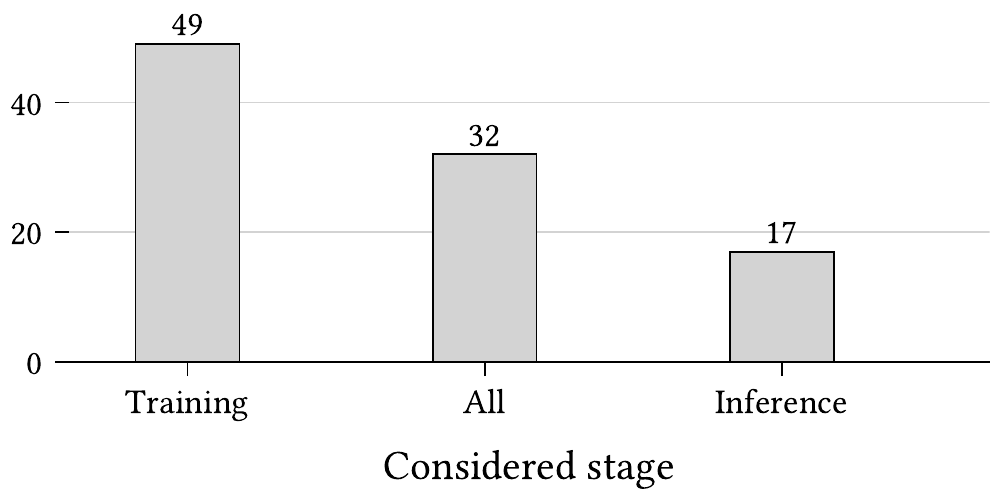}
    \caption{Number of publications per studied phase of AI.}
    \label{fig:res_phase}
\end{figure}

\begin{highlight}
\textbf{Green AI Pipeline Phase}\newline
\noindent \faLeaf~Approximately half of \ga studies focus on the training phase, while a minor portion considers the entire AI pipeline. Only a minor portion of the \ga literature focuses on the inference phase.    
\end{highlight}

\subsection{Considered Artifacts}
AI systems are based on several artifacts, and tackling the energy efficiency of such systems can thus involve multiple of those artifacts (\eg data, model, pipeline) or different related artifacts (\eg architecture, framework, CPU). A distribution of the artifacts considered in the primary studies is documented in Figure~\ref{fig:artifact}. The categories of artifacts are:

\begin{description}
    \item [Model] The publications within this category focus on the model and/or associated algorithm to tackle the energy efficiency of AI. \papercount{63}
    \item [Data] Papers that address energy efficiency through the study of the data used in the AI pipeline. \papercount{8}
    \item [Pipeline] Studies looking at the whole AI pipeline. \papercount{3}
    \item [Other] Publications dealing with CPU, architecture, and framework. \papercount{4}
    \item [General] The papers do not specify a particular artifact and address AI systems as a whole. \papercount{24}
\end{description}

\noindent{\begin{figure}[!h]
    \centering
    \includegraphics[width=\linewidth]{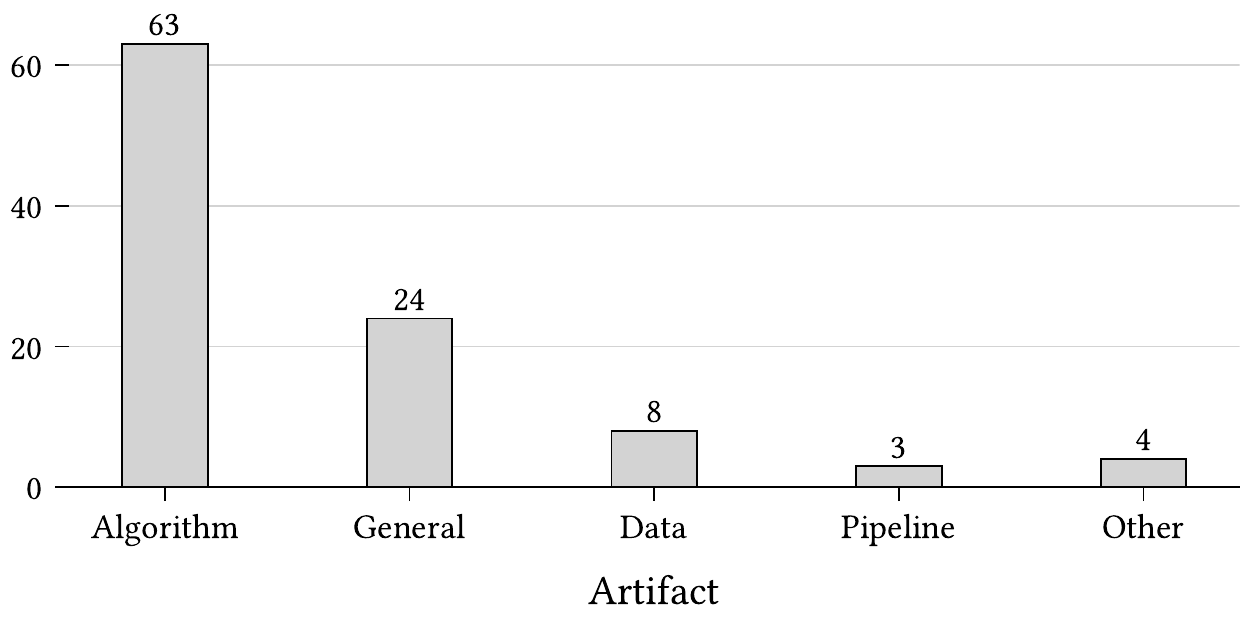}
    \caption{Number of publications per studied artifact.}
    \label{fig:artifact}
\end{figure}

\subsection{Algorithm Types}
By considering the primary studies which focus on a specific algorithm (\papercount{51}), we note that the vast majority focus on \textbf{neural networks} (\papercount{41}). Only a much smaller fraction focuses on algorithms of different nature, such as decision trees (\papercount{5}), genetic algorithms (\papercount{1}), or logistic regression models (\papercount{5}).

Regarding the deep neural network algorithms, we also note a further characterization of this field, with 8 studies focusing on \textbf{convolutional neural networks}, one on \textbf{transformers}, and one on \textbf{spiking neural networks}. We also observe three algorithms that appear only once in the Green AI literature (\textbf{Other} category, \papercount{3}), namely \textit{genetic algorithms}, \textit{logic regression algorithms}, and \textit{stochastic gradient descent algorithms}.

\begin{highlight}
\textbf{Green AI algorithm types}\newline
\noindent \faLeaf~Most \ga primary studies are algorithm-agnostic or focus on neural networks. A small fraction uses decision trees.
\end{highlight}

\subsection{Data Types Used}
Regarding the types of data used in the \ga body of literature, an overview of their distribution is reported in Figure~\ref{fig:data_types}. From the figure, we can observe that the recurrence of data types across primary studies is:
\begin{description}
    \item [Image data] \papercount{42} 
    \item [Textual data] \papercount{22} 
    \item [Numeric data] \papercount{10} 
    \item [Video data] \papercount{4} 
    \item [Audio data] \papercount{2}
\end{description}

From the distribution of data types, we notice that image data is by far the most used one, and is utilized by almost half of the studies in the body of literature. The second most utilized data type is textual data, which nevertheless appears approximately half as often as the image one. Other types of data result to be less recurrent, with only few studies utilizing audio data (\eg \citeauthor{lenherr2021Sep} present a metric to measure the sustainability of \ga by considering as case study the Intel MovidiusX processor, an embedded video processor with a Neural Engine for video processing and object detection~\citeP{lenherr2021Sep}).

A rather high number of primary studies does not specify any kind of data (\textbf{Not specified} category, \papercount{32}). This finding has to be primarily attributed to the position and theoretical papers included in the review (see also Section~\ref{sec:study_type} and Section~\ref{sec:topic}).

\begin{highlight}
\textbf{Green AI data types}\newline
\noindent \faLeaf~ Image data is the most used data type in \ga studies, followed by textual and numeric data.
\end{highlight}

\begin{figure}
    \centering
    \includegraphics[width=\linewidth]{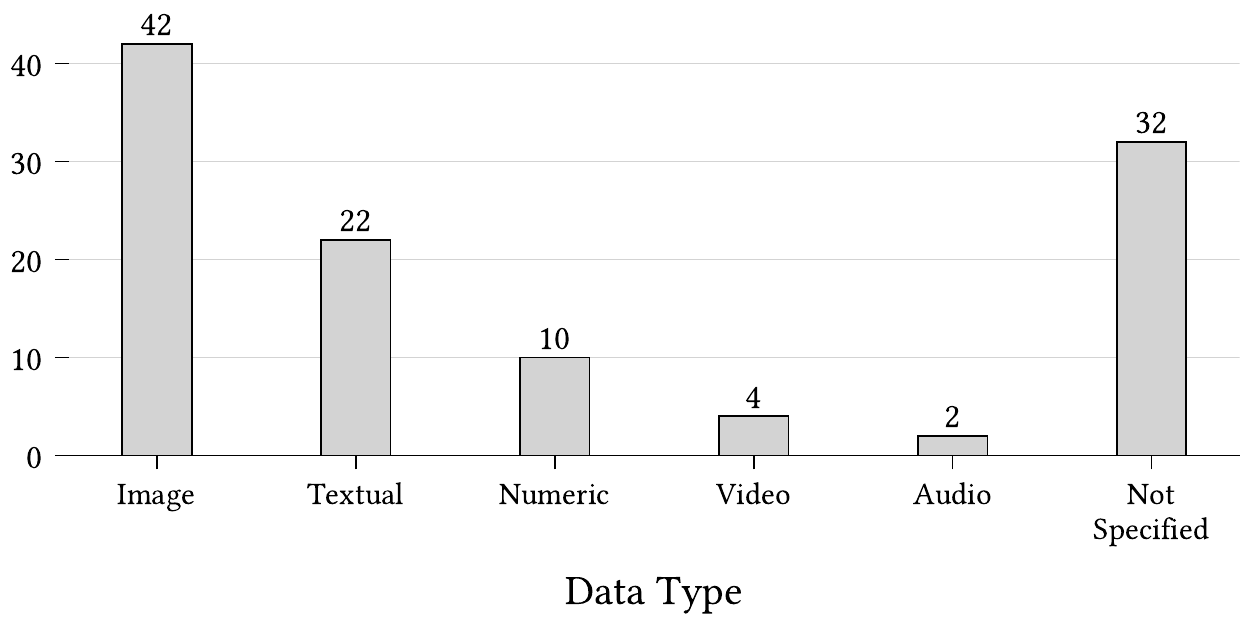}
    \caption{Occurrence of data types used in the \ga literature.\label{fig:data_types}}
\end{figure}

\subsection{Dataset sizes}
Regarding the size of the datasets used in the papers, approximately half of the primary studies (\papercount{48})) directly reference the number of data points used. By inspecting such numbers, we note that the number of data points used to study and to evaluate Green AI algorithms and approaches varies greatly, and ranges from \textbf{1k data points}~\citeP{Gondi2021Nov} to \textbf{40M data points}~\citeP{Garcia-Martin2017}. Almost half of the studies reporting the number of data points (\textit{$\triangleright$~\textbf{25} out of 48 papers}) utilize data points in the order of thousands ($1k\leq \# datapoints\leq 70k$), while the remaining (\textit{$\triangleright$~\textbf{23} out of 48 papers}) use one million data points or more ($1M\leq \# datapoints\leq 40M$).

\begin{highlight}
\textbf{Green AI dataset sizes}\newline
\noindent \faLeaf~ Dataset sizes range from 1k to 40M data points, with approximately half of the studies utilizing 1M or more data points.
\end{highlight}

\subsection{Research Strategies}
By considering the research strategies~\cite{stol2018abc} utilized in the \ga literature, the distribution of the various strategies, according to the collected primary studies, is reported in Figure~\ref{fig:validation_types}.

\begin{figure}
    \centering
    \includegraphics[width=\linewidth]{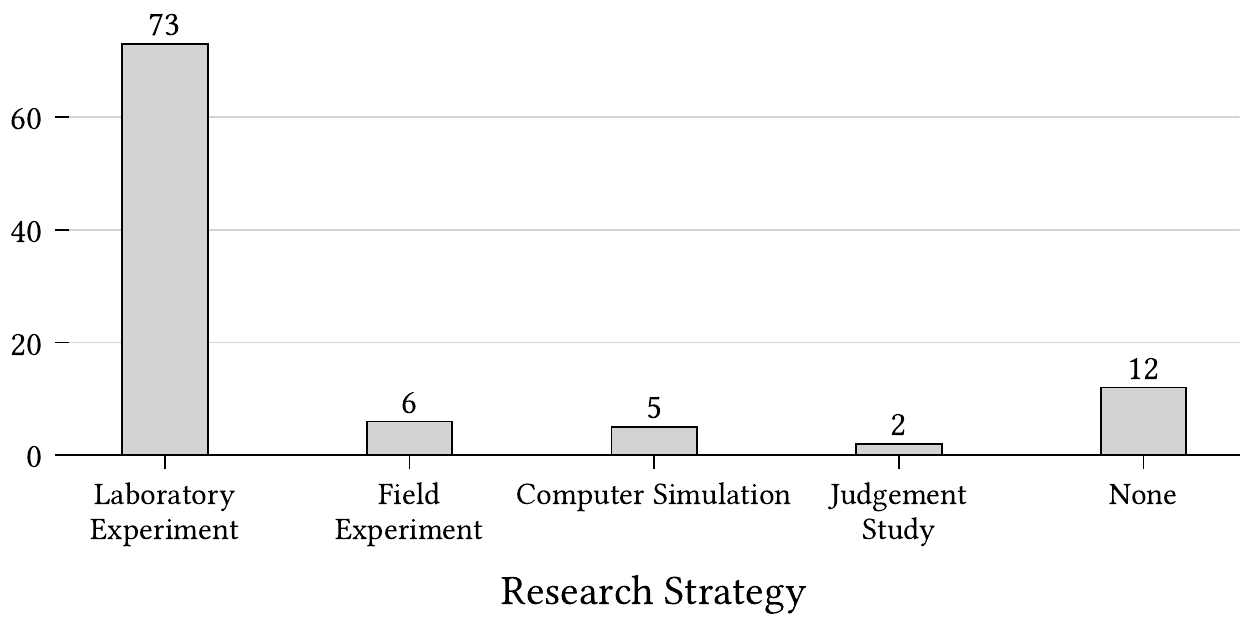}
    \caption{Occurrence of research strategies used in the \ga literature.\label{fig:validation_types}}
\end{figure}

The majority of paper results adopt \textbf{laboratory experiments} (\papercount{73}), while only a fraction uses other research strategies, such as \textbf{field experiments} (\papercount{6}), \ie experiments conducted in pre-existing settings and \textbf{computer simulations}, \ie ``in silico'' simulations conducted in a nonempirical setting (\papercount{5}). As examples, Liu~\etal~\citeP{liu2019} use a field study to assess a green software stack for computer vision of autonomous robots, while Yosuf~\etal~\citeP{yosuf2021} leverage computer simulations to study how virtualized cloud fog networks can be used to improve AI energy efficiency. \review{The 12 papers not displaying any research strategy correspond to the position papers (cf. the ``None'' category in Figure~\ref{fig:validation_types}).}

\begin{highlight}
\textbf{\ga Research Strategies}\newline
\noindent \faLeaf~ Most \ga studies use laboratory experiments, while only a minority adopt other research strategies, such as field experiments and computer simulations.
\end{highlight}

\subsection{Energy savings}
By considering the energy savings reported achievable \textit{via} Green AI strategies, we note that only approximately a third of the primary studies explicitly document them (\papercount{27}). Out of all Green AI strategies, among the ones which report concrete saving percentages, a technique based on structure simplification for deep neural networks results to save more energy, amounting to \textbf{115\% energy savings}~\citeP{Zhang2018}. The other techniques which result to optimize energy the most are based on quantizing the inputs of decision trees~\citeP{abreu2020} (97\% energy savings), using data-centric \ga techniques~\citeP{Verdecchia2022DataCentricGreenAI} (92\% energy savings), and leveraging efficient deployment of AI algorithms \textit{via} virtualized cloud fog networks (91\% energy savings)~\citeP{yosuf2021}. Overall, more than half of the papers explicitly reporting energy saving percentages report a saving of at least 50\% (\textit{$\triangleright$~\textbf{17} out of 27 papers}), while only a minor number savings between 13\% and 49\%.

\begin{highlight}
\faLeaf~\textbf{Green AI energy savings}\newline Studies report energy savings between 13\% and 115\% energy savings, with more than half of the papers reporting savings of at least 50\%. 
\end{highlight}

\subsection{Industry involvement}
Regarding the industry involvement in \ga scientific publications (see also Section~\ref{sec:data_extraction}), an overview of the authorship of the \ga primary papers is depicted in Figure~\ref{fig:euler}.

\begin{figure}
    \centering
    \includegraphics[width=0.8\linewidth]{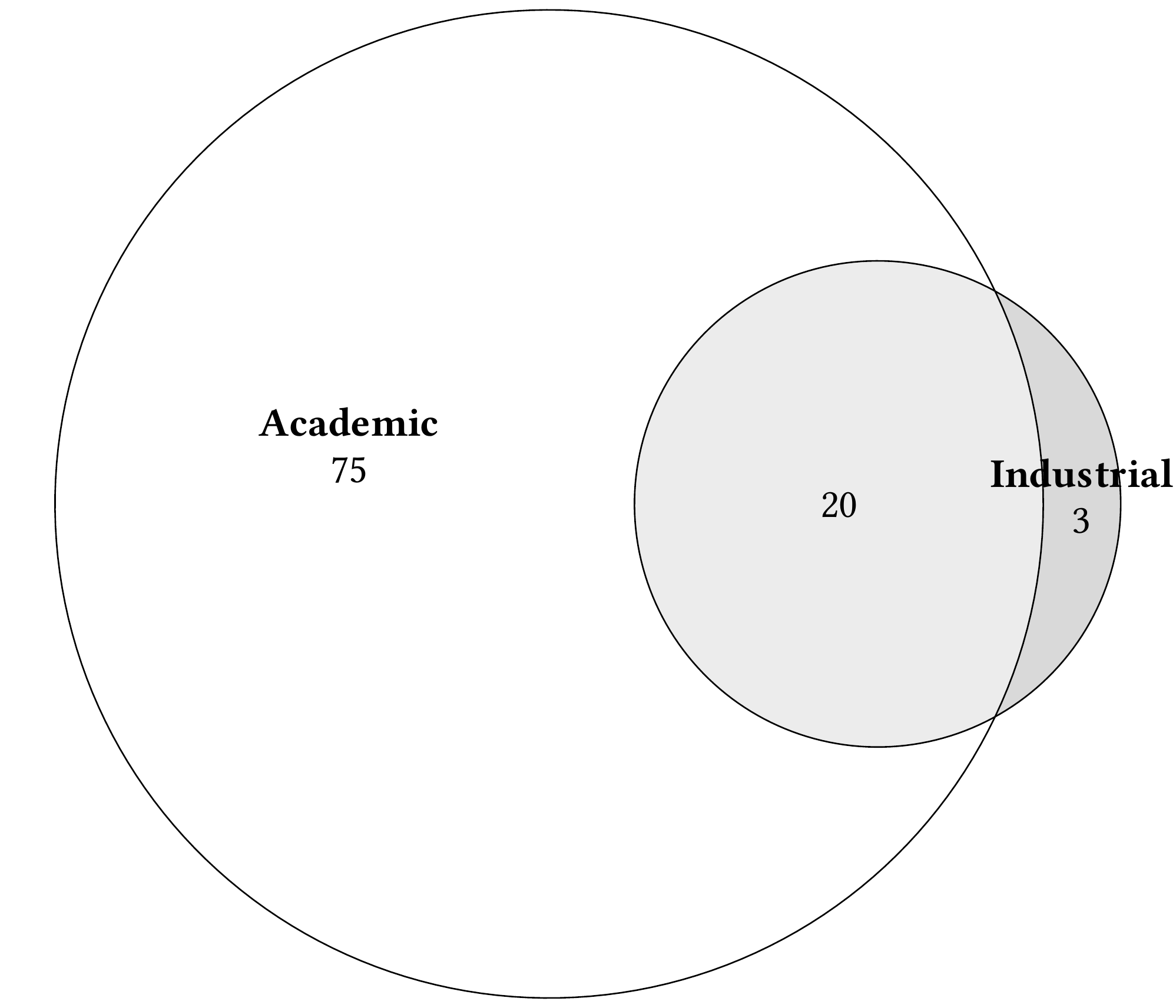}
    \caption{Industry involvement.\label{fig:euler}}
\end{figure}

From the figure, we can note that most \ga studies are authored exclusively by academic researchers (\papercount{75}), while also a considerable portion, amounting almost to a fourth of all primary studies, are authored by a mix of academic and industrial researchers (\papercount{20}). \ga studies written exclusively by industrial authors appear only in rare instances (\papercount{3}).
\begin{highlight}
\faLeaf~\textbf{Industry involvement}\newline 
Most studies are written by academic authors, while a minor portion by a mix of academic and industrial authors. \ga studies written exclusively by academic authors are very rare.
\end{highlight}

\subsection{Intended readers}
By considering the intended readers of the \ga scientific literature, the vast majority targets academic readers (\papercount{85}), while a much smaller portion both academic and industrial readers (\papercount{8}). Despite scientific papers targetting intuitively a specialized audience, among the \ga literature, few studies are intended also for the general public (\papercount{5}). For example, Dhar~\etal~\citeP{dhar2020}, present an intuitive yet thoroughly positioned article on the systemic effect of AI on carbon emissions. Interestingly, among the primary studies, few are intended also for policymakers, \ie aim to sensibilize government stakeholders to consider issues related to \ga. For example, in a paper by Rohde~\etal~\citeP{rohde2021}, how opportunities and risks for the environment, economy and society associated with AI can be governed are discussed. 

\begin{highlight}
\faLeaf~\textbf{Intended readers}\newline 
The vast majority of \ga studies are targetting academic readers, while a much smaller portion targets both academic and industrial readers. A handful of studies, especially position papers, are intended for the general public.
\end{highlight}

\subsection{Tool Provision}
Among the primary studies collected for this literature review on \ga, only a small fraction (\papercount{15}) makes tools available to tackle \ga. The tools provided are of heterogeneous nature, and range from tools to monitor the resource efficiency of AI algorithms~\citeP{guldner2021}, to tools optimizing the energy efficiency for stochastic edge inference~\citeP{kim2020}, and implementations of convolutional neural networks optimized for energy efficiency~\citeP{mehta2019}.

\begin{highlight}
\faLeaf~\textbf{\ga Tool Provision}\newline 
Albeit numerous studies provide solution to tackle \ga, only a fraction of them makes tools based on the solutions readily available online as an implemented tool.   
\end{highlight}

\section{Discussion}\label{sec:discussion}

\textbf{\textit{\review{The consolidated and still growing \ga publication trend.}}} From the analysis of the publication trends a clear picture emerges. The topic is gaining increasing traction in the academic community, especially if the latest years are considered (from 2020 onward). Despite being a quite new research topic (with the first paper on \ga being published in 2015), the socio-environmental relevance of the topic seems to be reflected in its targeted publication venues. With conferences and journal being the most recurrent \ga publication venues, the \ga research field seems to have positioned and consolidated itself quite quickly within AI research communities.

\review{\textbf{\textit{A definition of \ga.}} From the results regarding how the term ``\ga'' is used in the literature a clear picture emerges. Most \ga studies consider \ga as exclusively related to energy efficiency. Only fewer studies examine the influence of AI on greenhouse gas emissions ($CO_2$), and an even minor fraction examines the holistic impact that AI has on the natural environment.}

\review{By considering the different levels of abstraction (namely \textit{energy efficiency}, \textit{carbon footprint}, and \textit{environmental footprint}) the higher, more encompassing level, of environmental footprint seems best fitted to define the field of Green AI. In fact, as demonstrated in recent literature, reducing the environmental impact of AI exclusively to energy consumption has to be deemed as overly simplistic process~\cite{luccioni2023counting}. Similarly, as green resources are sustainable but not infinite~\cite{verdecchia2022future}, the field of Green AI has to account also for the multifaceted environmental impact AI can have, other than $CO_2$ emissions alone. Based on these considerations, we define the field of \ga as follows:}

\begin{nbquote}
    \review{\textit{``\ga regards practices aimed at utilizing AI to mitigate the impact that humans have on the natural environment in terms of natural resources utilized, and/or mitigating the impact that AI itself can have on the natural environment.''}}
\end{nbquote}

\review{On one hand, the definition above perfectly fits the studies focusing on the holistic impact that \ga has on the natural environment. On the other hand, given its encompassing nature, the definition is also suited for studies focusing on lower abstraction levels of sustainability, such as \ga $CO_2$ emissions and energy consumption. In the latter case however, the definition also acts as a word of warning: while studying the lower levels of \ga is paramount, only by considering the totality of the heterogeneous natural resources utilized by AI can we really understand the environmental impact of AI.}

\textbf{\textit{\review{The transdisciplinary topics of \ga (with gaps).}}} The 13 different topics we discover in this review emphasize that Green AI is a broad field that needs to be tackled as a transdisciplinary field. Some topics are naturally tied to training strategies (\eg monitoring, hyperparameter tuning, algorithm design). However, there are other topics that take Green AI outside the training realm.

This is the case for example of Deployment, Libraries, and Estimation that promise to be relevant in enabling Green AI. We argue that other disciplines need to be involved. For example, Software Engineering which has been dealing with these topics for traditional software systems. As highlighted by Cao~\etal in their work on estimation~\citeP{Cao}, one cannot expect existing strategies for traditional software to address the new challenges of AI-based systems. Conversely, only a few Green AI papers~\citeP{Georgiou2022May,McIntosh2019Mobile,Hampau2022} come from software engineering venues. 

Our analysis also shows that the topics Estimation and Emissions are under-represented, with six and five papers, respectively. We argue that more work is quintessential in these topics to help scientists and practitioners report the carbon footprint of their AI models in a seamless way.


We showcase that papers under the topic Policy are only covered by position papers. We find this finding disconcerting: new policies to encourage Green AI within both industry and academic contexts need to be backed up with reliable evidence. Hence, we need more observational and solution papers that tackle this topic in the near future.

The same issue is present in Emissions -- only one paper is observational and the remaining are position. It might be the case that computing the climate impact of AI is far from trivial and it is easier said then done. Again, this is a call for the community to take action. It is not enough to ask big companies to provide their data on carbon impact -- we also need to provide strategies and solutions to make it standard and straightforward.

\textbf{\textit{\review{The fundamental \ga research unbounded from application domains.}}} From the collected results we deduce that, in order to improve the environmental sustainability of AI, it is often not necessary to focus on a specific domain. This implies that frequently fundamental aspects of \ga are still open to investigation, and results can then be ported from a generic setting to specific domains. However, from the obtained results, we also note that the increasing distribution of digital infrastructures to achieve environmental sustainability~\cite{verdecchia2022future} might have played a role in \ga research, with edge computing being the most considered specific domain.

\textbf{\textit{\review{The high emphasis on the AI training phase.}}} The results regarding the AI pipeline phases considered in the literature unequivocally point to training as the most studied phase. Albeit the training phase is intuitively the most energy-greedy phase, this results calls for a word of caution. From recent results (\eg a study on data-centric \ga~\citeP{Verdecchia2022DataCentricGreenAI}) the inference phase results to consume only a negligible fraction of the energy consumed in the training phase. Nevertheless, given the high execution rate of the inference phase, how the energy consumed by the infrequent execution of the training phase compares to one of the highly executed inference phase is still an open question. As a call for action, studies should be conducted by considering the energy consumed throughout the whole life cycle of AI models, from their training to inference phase, till their eventual deprecation. 

\textbf{\textit{\review{Image datasets as primary \ga data source}.}} By considering the data types used in \ga studies, we note that the vast majority of the literature uses image data. To the best of our knowledge, this choice is not guided by any specific research design choice (\eg AI models based on image data being the most used in practice, or being the most energy greedy ones). For this reason, we conjecture that the popularity of utilizing image data for \ga data is mostly driven by convenience, either because past work focused on such data by chance, image datasets are more accessible/standardized with respect to other ones, or more off the shelf image AI models/libraries are currently available. Regardless of the cause, this result points to the need of utilizing more heterogeneous data types, rather than focusing primarily on image data, in order to gain a holistic understanding of \ga.

\textbf{\textit{\review{Laboratory experiments guided till now \ga.}}} The most common research strategy adopted for \ga studies clearly emerges from the literature as being laboratory experiments. Given the fast popularization and consolidation of the \ga research field, from this review it seems as if the time is suitable to shift the focus to other research strategies, \eg field experiments and case studies. This would not only allow to change the considered context from an \textit{in vitro} to an \textit{in vivo} setting, but also to bridge potential gaps between academic research and industrial practice.

\textbf{\textit{\review{The highly promising energy savings of \ga.}}} From the results of this review, we deduce that the research field of \ga is highly promising, with more than half of the papers reporting 50\% or more energy savings. This study focuses on the state of the art of \ga, rather than focusing on the state of practice. It would be therefore interesting to understand, as future work, the extent to which this encouraging results are transposed to industrial practice, and the potential impediments which hinder their adoption or full potential.

\textbf{\textit{\review{A noticeable industry involvement}.}} Regarding industry involvement in \ga studies, the results gathered from this review are promising. The authorship of \ga literature results to be to a good extent shared between academic and industrial researchers/practitioners. This finding might highlight the sensibility of industry towards \ga concerns, and/or the importance of moving towards more environmentally sustainable AI practices.

As a potential impediment to the industrial adoption of \ga research, our results point to a low recurrence of studies targeted towards practitioners. While numerous journals are explicitly aimed at practitioners, \eg IEEE Software\footnote{\url{https://www.computer.org/csdl/magazine/so}. Accessed 22nd December 2022.}, only few studies on \ga included in our review target them. This result might point to the fact that the \ga interest is still primarily focused towards academic activities, while the authorship showcases a rather high interest of industry. As take away, similar to the considerations made for the \ga research strategies, it might be the right moment to consider a higher involvement of industry in \ga, which results to date to be a research area still targeted primarily towards academic readers.

\textbf{\textit{\review{\ga lacks tool support}.}} Finally, from this review, we note that the current situation regarding the provisioning of \ga tools is not bright. Albeit the majority of the studies present \ga solutions, only a small fraction of them makes the solutions available as a tool. We conjecture that this result might either point towards (i) a fast-paced nature of \ga research, in which results are rapidly deprecated, and hence tools are not meaningful, or (ii) an immaturity of the research field, which still requires a solid empirical foundation on which tools can be~built~upon.

\section{Threats to Validity}
\label{sec:threats}
In this section, we discuss the threats to validity of our study. 
To ensure the quality of the results, we established a well-defined research protocol to proceed with the data collection. In addition, throughout our study, we followed the recommendations of the guidelines for conducting a systematic literature review~\cite{Kitchenham2004, Wohlin2014, mayring2004qualitative, kitchenham2013systematic, petersen2015guidelines}. We designed and carried the different reviewing processes according to the rigorous protocol we established after the guidelines and described in Section~\ref{sec:method}. Nevertheless, some threats to validity can still exist even with our best efforts. In the following, we present the threats which could have influenced our study, jointly with the strategies we adopted to mitigate them.

\textbf{External validity}
The main threat to external validity is that the literature collected and analysed in this study is not sufficiently representative. To avoid this situation, we surveyed three prominent literature indexers through an automatic query (\ie Google Scholar, Scopus, and Web of Science), and left the year of publication unbounded, to reduce the probability of missing any relevant publication. In addition, the search query was designed to target relevant literature directly with specific keywords, while allow for flexibility by considering similar, complementary, and variation of the keywords (\eg the keywords \textit{green}, \textit{sustainability}, and \textit{sustainable}). We also mitigated the threat of having an incomplete set of studies, as well as the threat associated with the specificity of the terms used in the search query, by performing a complementary iterative bidirectional snowballing process of the query results. This latter search strategy allowed us to include literature related to our query that was not directly referencing any of the automated search keywords. 
We limited our review of the literature to peer-reviewed studies, to moderate the threat about the low quality of the set of primary studies. We deem that such practice does not constitute an additional threat, as peer-review is a standard requirement of high-quality publications.

\textbf{Internal validity}
To address potential threats to internal validity, we established a rigorous research protocol \textit{a priori}, and we followed it to conduct all the research activities. Subjective biases and interpretations were mitigated by closely complying with the selection criteria to evaluate the studies.
Moreover, weekly meeting were held during the selection process to jointly discuss examples, doubts, and to align the selection process between the three researchers.

\textbf{Construct validity}
To ensure that the set of studies answered our research questions, we applied \textit{a priori} carefully constructed inclusion and exclusion criteria to strictly control the manual selection of studies. We then used the bidirectional snowballing technique to expand the range of relevant primary studies to a more comprehensive set.

\textbf{Conclusion validity}
Possible sources of bias arising from the data extraction and analysis phases were mitigated by strict compliance with an \textit{a priori} defined protocol, explicitly tailored to collect the data needed to answer our research questions. In all, we followed the best practises of the standard guidelines for systematic literature reviews~\cite{Kitchenham2004, Wohlin2014, mayring2004qualitative, kitchenham2013systematic, petersen2015guidelines}.
Lastly, we documented all the data throughout the whole review process and made them available for reproducibility and replicability purposes (see Section~\ref{sec:intro}).

\section{Related Work}\label{sec:rw}
Despite the growing interest around Green AI, the topic has been marginally considered only in a handful of reviews. The related work manly investigates the topic as an intersection of AI and environmental sustainability, or by defining it as a specific subdomain of software engineering. To the best of our knowledge, this review is the first aiming towards a comprehensive review of  Green AI research and its characteristics. 

In a recent publication, Natarajan~\etal perform a systematic literature review on the topics of `AI for Environmental Sustainability' as well as `Environmental Sustainability of AI’. The authors present the affordances of the use of AI for sustainability that they extracted from the literature~\cite{Natarajan2022}. `AI affordances' are introduced as the posible actions offered by AI artifacts to an organizational actor whose goal is to achieve environmental sustainability. The authors point out the focus of previous research on the technical side, and they advocate for a further exploration of the concept of sustainable AI affordances from a socio-technical perspective. The literature is exclusively analyzed with respect to building the AI affordances, and other characteristics of the state-of-the art of \ga are considered nor discussed in the study.
In contrast, our review focuses on the sustainability of AI, and maps the entirety of the Green AI literature. In our review, we aim at providing a detailed and comprehensive overview of the characteristics of the \ga state-of-the-art research (\eg topic, domain, type of study, targeted artifact, overview of energy savings, tool provision, industrial involvement). Therefore, in contrast to the work of Natarajan~\etal~\cite{Natarajan2022}, we consider the different facets of \ga, rather than exclusively on AI affordances, leading to a more holistic review of \ga, and a higher number of considered primary studies (98 versus 41 papers).
This difference could be explained by the fact that their review only includes papers involving consumer products and services and excludes papers dealing with non-commercial applications, whereas we provide an overview of the whole field of \ga.

Previous literature reviews consider \ga research by focusing exclusively on specific subdomains of AI and application subdomains of Software Engineering, \eg deep learning~\cite{Xu2021}, information retrieval~\cite{Scells2022}, or embedded systems~\cite{Mittal2019}. In contrast, our research aims to review the entirety of the \ga literature, regardless of the specific AI or software engineering subdomain it focuses on.

In the survey of \citeauthor{Xu2021}~\cite{Xu2021}, the authors provide an overview of the approaches aimed at improving the environmental sustainability of deep learning. The authors map the different approaches using a taxonomy of the deep learning life cycle stage and its related artifacts. In contrast to such study, in this review we target a higher number of \ga characteristics (see Section~\ref{sec:data_extraction}), and target the entirety of \ga literature, rather than exclusively the one on deep learning. 


\citeauthor{Scells2022}~\cite{Scells2022} provide a literature review on methods related to the domain of Green Information Retrieval. The authors explain that the domain of Information Retrieval (IR) produces relatively low emissions compared to other research domains, but they also warn that similar trends of costs and environmental impact may appear considering the growing development of new IR-focused deep learning models.
Natural Language Processing and Machine Learning are also discussed, but only with respect to the Information Retrieval domain. Therefore, they are not addressing the whole field of AI, as done in this review.

Finally, the optimizations that can be made for the implementation of deep learning models on the specific platform of NVIDIA Jetson are reviewed with a focus on energy efficiency by \citeauthor{Mittal2019}~\cite{Mittal2019}. The review covers studies at both the hardware and software level. Nevertheless, the review addresses only the Jetson platform~\footnote{\url{https://developer.nvidia.com/embedded-computing}. Accessed 23th December 2022.}. We differentiate ourselves from this study by providing a holistic review of Green AI, rather than focusing exclusively on deep learning.

\section{Conclusion}
\label{sec:conclusion}

In this systematic literature review, we aimed at characterizing the existing body of research in Green AI. We identified 98 peer-reviewed publications that show a significant growth in this research field since 2020.

We provide an encompassing overview and characterization of the different topics being addressed by \ga papers. We identified 13 different \ga topics, showcasing that the spotlight falls on monitoring, hyperparameter-tuning, model benchmarking, and deployment. Less frequent topics -- such as data-centric, estimation, and emissions -- show less obvious approaches that deserve further research in the upcoming years.

The potential of Green AI cannot be disregarded: the majority of publications show significant energy savings, up to 115\%, at little or no cost in accuracy. However, we argue that most publications revolve around laboratory studies. More field experiments are quintessential to help AI practitioners embrace green strategies that are effective, feasible, and mensurable. This is also reflected in the small participation of the industry in these studies -- only 23\% of publications involve industry partners. 

At the same time, we conclude that the field seems to be reaching a considerable level of maturity. Hence, it is necessary to encourage the port of promising academic results to industrial practice.
In other words, our study calls out for the importance of having reproducible research. Only a small fraction of solution papers offers a tool or software package that can be used by the community. We argue that Green AI is an urgent and necessary line of research that needs to grow fast and solid -- non-replicable research can only slow us down.

This review also serves as a foundation for future research that ultimately aims to reduce the climate impact of AI. In this respect, we see potential in follow-up grey literature or interview studies to understand how AI professionals are currently addressing the issue.

\balance
\bibliographystyle{ACM-Reference-Format}
\bibliography{references}

\nociteP{*}
\bibliographystyleP{ACM-Reference-Format}
\bibliographyP{primaries}



\end{document}